# A Comparison of Neural Network Training Methods for Text Classification

**Anderson de Andrade**
Department of Computer Science
University of Toronto
Toronto, ON M5S 2E4
*adeandrade@cs.toronto.edu*

## Abstract

We study the impact of neural networks in text classification. Our focus is on training deep neural networks with proper weight initialization and greedy layer-wise pretraining. Results are compared with 1-layer neural networks and Support Vector Machines. We work with a dataset of labeled messages from the Twitter microblogging service and aim to predict weather conditions. A feature extraction procedure specific for the task is proposed, which applies dimensionality reduction using Latent Semantic Analysis. Our results show that neural networks outperform Support Vector Machines with Gaussian kernels, noticing performance gains from introducing additional hidden layers with non-linearities. The impact of using Nesterov's Accelerated Gradient in backpropagation is also studied. We conclude that deep neural networks are a reasonable approach for text classification and propose further ideas to improve performance.

## 1 Introduction

In the context of neural networks on image classification tasks, there is an impact in introducing additional hidden layers with non-linearities, with examples in [1]-[3], and many other research endeavors. Every upper layer introduces a higher level of abstraction that is very successful in predicting the label of objects, shapes and patterns in images. We want to see if deep neural networks are also able to build interesting high level features for text classification.

Extracting domain-specific information from new data sources has many useful applications in increasing demand. The dataset employed corresponds to messages from the Twitter microblogging service, provided by the CrowdFlower Open Data Library [5]. The purpose of the learning algorithm is to predict several characteristics of weather-related messages, including sentiment, time of occurrence and type.

We hypothesize that this dataset can benefit from the use of deep neural networks as they will be able to capture correlations in data to generate more complex features required for these tasks, specially sentiment analysis, which usually requires higher levels of understanding.

Stacking hidden layers with non-linear activation functions introduces difficulties during training. As stated in [2] and [4], at the beginning of learning, units in the upper layers will saturate and prevent gradients to flow backward, stopping learning in the lower layers. Eventually, after many epochs, the units will leave saturation and training of lower layers will begin again, but this can take a long time and require large quantities of labeled data. With proper initialization or pretraining of the weights, and careful selection of activation functions, the process can be sped up.

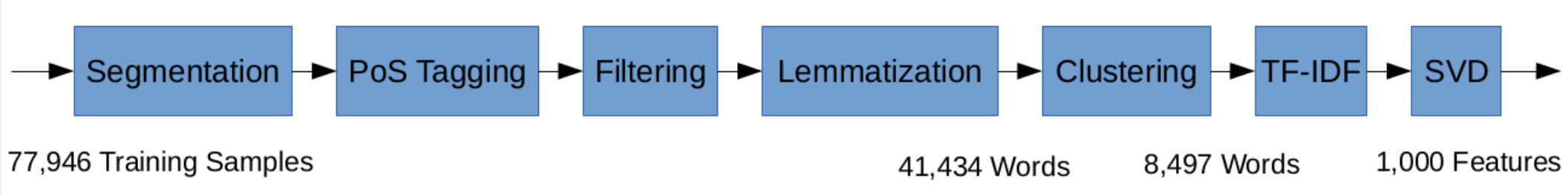

Figure 1: NLP modules composing the feature extraction pipeline. For 77,946 training samples, 41,434 token-tag pairs are obtained. After clustering, they are trimmed down to 8,497. Finally, singular value decomposition reduces feature vectors to 1,000 dimensions.

Although unsupervised pretraining by Deep Belief Networks (DBN) and Deep Boltzmann Machines (DBM) can still be considered the state of the art in training deep neural networks, the task is still computationally expensive and requires fine-tuning of the final classifier. More recently, it was discovered that deep supervised networks can be trained by proper weight initialization. It was empirically demonstrated by Gloriot and Bengio in [2] that making weights large enough allows gradients to flow well and activations to convey useful information.

Another interesting ingredient is the use of momentum [4] to speed gradient descent in deep networks. Other non-linear activation functions such as Maxout [1] or Softsign [2] have been very successful in the same task. These purely supervised methods for training deep neural networks have offered good results and seem to perform specially well with large quantities of data [1].

The scope of this project is to the implement proper initialization [2] with tanh activation functions, and greedy layer-wise pretraining of a DBN using RBMs [3]. We will compare how these two techniques fare using Nesterov's Accelerated Gradient (NAG) as in [4].

## 2 Methodology

### 2.1 Feature extraction

The selected dataset has a fair amount of non-standard English, which results in a larger feature space, making clustering difficult because it requires the use of either specialized dictionaries, which may not cover cases out of its scope, or large dictionaries trained with unsupervised approaches, that may lack precision.

The short property of the sentences in the dataset makes the feature vector very sparse. As a result, even a small dimensionality reduction can lose much of the information in a training sample, affecting its prediction by the machine learning algorithm and introducing noise during the training phase.

Thus, selecting relevant features and deciding how to encode them will have an enormous impact on the learning method's ability to extract a good model. An algorithm composing a processing pipeline of small NLP modules was designed as shown in Figure 1. It contains the following steps:

1. Word lengthening is commonly observed in microblogging; strip down characters repeated 4 or more times to 3 (i.e. *hooooottttt* is converted to *hoootttt*). The resulting word is able to retain the intended heightened emotion that can help sentiment analysis.

2. Tokenize messages using [6], which correctly splits multiple sentences and includes patterns for common non-standard English behaviors, such as compound words, additional punctuation marks and *emoticons*.

3. Tag every token with a part of speech using [6], configured with a model that uses the Penn Treebank set extended with additional tags to support Twitter messaging functions and metadata, such as *mentions*, *retweets* and *hashtags*. This tag set is specific enough to provide important information such as verb conjugation and noun inflection.

4. Metadata *hashtags* are segmented using [7]. The procedure involves calculating the most probable segmentations, using the accumulated probabilities of bigrams collected from the Microsoft Web N-gram Service. The results are then ran through the same procedures of step 2 and 3 and the token-tag pairs are incorporated in the parent data structure.

5. Split compound words concatenated with dashes or slashes, assign them the same tag of the original word and, remove the parent pair and inject the new pairs.

6. Remove specific stop words and part-of-speech tags considered not useful in the classification task: determiners, particles, conjunctions, pronouns and all punctuation except for exclamation and question marks, which are potentially useful in sentiment analysis.

7. Lowercase and lemmatize every token, employing the WordNet lemmatizer provided in NLTK [8], according to [9].

8. Generate a bag of words from token-tag pairs, tracking the number of occurrences among all training examples. Separate identical words by tag according to the following: a tag for nouns and proper nouns, a tag for adjectives, a tag for adverbs, and a different tag according to different tenses of the verb: base form, past tense, gerund or present participle, and past participle; identical tokens under different tags are grouped together.

9. Prune bag of words according to a count threshold *K*. For every token-tag pair that did not occur in the training samples more than *K* times, perform a breadth-first search in WordNet's hyponymy hierarchy, using [8] according to [9]. The first lemma found in the bag with more than *K* counts gets added to a conversion list for later use. Words not found in the dictionary or with synonyms or hypernyms not found in the bag of words are dropped.

10. Generate a binary feature vector from the resultant bag of words. If a token-tag pair is found in the conversion list, the corresponding feature gets assigned.

The lemmatization and pruning steps in the algorithm allows to cluster words accurately and reduce dimensionality. However, they rely in dictionaries that do not support non-standard English lemmas and so many tokens won't have associations. We considered larger dictionaries such as *word2vec* [10] but after further evaluation it was found that the closest words obtained may belong to different senses that the one intended, an issue usually avoided in WordNet when transversing according to the most common senses in a breadth-first manner. Note that although using a stemmer in step 7 could cluster more features together, well-formed lemmas are needed to make use of a dictionary in step 9.

Since sentences in the training samples are short, it is unlikely for a non-stop word or tag to occur more than once. Working with binary feature vectors can help the algorithm be computational tractable while still representing the dataset well enough. Keeping verb tags that specify tenses could be useful to predict the time of occurrence in the classification problem.

Note also that although there may exist many words with spelling errors, which are not considered non-standard English, an attempt to fix them can bring noise to the dataset; the recall of popular spellchecking libraries is often very high, and their precision rather low, even when introducing some context with bigrams and trigrams. This is due to the design choice of these packages to make suggestions even when their probabilities are low, relying on user input to make the final decision.

For the 77,946 training samples processed, a bag of words with 41,434 token-tag pairs was obtained. After applying the clustering procedure in step 9 with $K = 5$, the number of features was reduced to 8,487.

### 2.2 Dimensionality reduction

To keep the problem tractable, we need to reduce dimensionality even more. This could be achieved by raising the threshold *K* in our clustering procedure until we get a reasonable number of features, but since the drop rate of words is high (79.62% of the words below the threshold $K = 5$), it would be better to compress features by projecting them onto lower dimensions that still capture the manifold created by the data.

The procedure would also get rid of dimensions with strong correlations, something that clustering cannot achieve. Since the dataset has a large number of data cases and dimensions, we decided to use singular value decomposition (SVD), which operates on the feature vectors directly, instead of on a covariance matrix.

A sensible approach before performing this step, was to weight the feature vectors with the Term Frequency-Inverse Document Frequency (TF-IDF) score, so it reflects how important a token-tag pair is to a message in the dataset that hopefully represents the population. This step will also effectively remove stop words that were not manually selected in our feature extraction implementation. The use of TF-IDF and SVD together is frequently referred as Latent Semantic Analysis (LSA). When performed the mentioned transformations using [11], the number of features was reduced to 1,000, a reasonable quantity that will keep the rest of the computations tractable.

### 2.3 Model design

The following models were considered:

1. A feed-forward neural network of 1 hidden layer with sigmoid activation functions and random weight initialization obtained by sampling from a Gaussian distribution with mean zero and variance 0.01.

2. A deep neural network with 3 hidden layers, tanh activation functions, and a normalized weight initialization as proposed by [2].

3. A deep neural network with 3 hidden layers, sigmoid activation functions and weight pretraining with Deep Belief Networks (DBN) as described in [3].

4. A set of support vector machines with non-linear Gaussian kernel functions (RBF).

The 1-layer neural network will provide us with a reliable baseline for performance analysis of our deep neural networks, since increasing the number of hidden layers with non-linear units will prove optimization more unstable due to unit saturation, where their gradients propagated backwards into the lower layers might not be sufficient to move the parameters into regions corresponding to good solutions, hypothesizing that the lower level parameters got stuck in a poor local minimum or plateau.

It was hypothesized in [2] that the transformation the lower layers of the randomly initialized network initially computes is not useful to the classification task, making the lower layers rely more on its biases, the error gradients pushing the activations towards 0. Pushing sigmoid outputs to 0 will bring them to saturation, preventing gradients to flow backward and stopping learning in the lower layers. In the case of symmetric activation functions though, sitting around 0 is good because it allows gradients to flow backwards.

In the same study, it was also presented that just after initialization, the back-propagated gradients are smaller and their variance decreases as they move to the lower layers. A new proposed initialization procedure approximates the optimal value of an objective function that maintains the variance of activations and back-propagated gradients as we move up or down the network. Deep neural networks using hyperbolic tangent activation functions with this normalized initialization procedure showed promising results that we will try to replicate.

In our third model, weights are pretrained with unsupervised Deep Belief Networks. Since the inputs of the model after performing dimensionality reduction are no longer binary but real valued, two different models of RBMs are used: in the first, real-valued visible units are connected to binary hidden units. In the second, binary visible units are connected to binary hidden unites. The former model is used to pretrain the weights between the input layer and the first hidden layer, the latter trains the adjacent hidden layers. The pretraining did not use any information about the class labels. After computing the weights, we connected the output units to the top layer, set the weights of that layer using normalized initialization and fine-tuned the whole network.

All hidden layers have 1,000 units, same number as the dimensions of the input. Every hidden layer was designed to have the same number of units mostly because the normalized initialization method requires so. As for the output, the weather classification problem involves 24 classes, grouped in different categories expressing: sentiment with 1-of-5 classes, time of occurrence with 1-of-4, and type of weather with N-of-15 classes where $N \leq 15$. This gives us a total of 17 classifiers, 2 of them have multiple classes and 15 are binary. The output layer was modeled with 24 units, the multi-class classifiers are grouped with the softmax function and the binary classifiers have sigmoid activation functions.

Support vector machines offers a robust, stable solution that will measure the feasibility of using neural networks to solve this particular problem. It won't be used as a baseline though, since we are mostly interested in the impact of deep neural networks in modeling the data. An off-the-shelf implementation from [11] was used, creating a model for each of the 17 classifiers.

### 2.4 Learning

Having 77,946 labeled samples, it is reasonable to split them into 90% training data and 10% validation data. Learning of the 1-layer neural network and deep neural network with normalized initialization is done with the backpropagation algorithm. Fine-tuning of the DBN also uses backpropagation. Optimization of the neural networks was ultimately performed using Nesterov's Accelerated Gradient (NAG) according to [4]. Mini-batches of size 100 were used. In the single-layer neural network, the momentum coefficient was initially set to 0.5 for the first 5 epochs and then raised to 0.9 for the subsequent epochs. With the deep neural networks, using a high momentum produced unstable results, so it was kept fixed at 0.5. The learning rate was set to 0.12.

Weight decay was not used in favor of early stopping, in which the best version of the neural network on the validation set is kept while tracking increases of the error rate. The training will stop when the error increases in 2 successive epochs, returning the best version of the neural network found.

The targets in the dataset were manually labeled by multiple raters, and some amount of disagreement was expected. The mixture of labels that the raters gave a training case and the individual trust of each rater were used to generate confidence scores or soft probabilities. Instead of converting the scores to hard probabilities and using the cross-entropy error function, we rely on the root mean square error (RMSE). For in-depth analysis the classification error rate of every classifier will be computed by converting the targets and predictions to hard probabilities and comparing.

In our DBN, greedy layer-wise pretraining was used according to [3]. The RBMs having binary visible and hidden units were trained for 50 epochs with a 0.1 learning rate. Pretraining the first layer of features required a much smaller learning rate to avoid oscillations; the learning rate was set to 0.001 and pretraining proceeded for 200 epochs, ensuring a stationary distribution is reached. For both models the regularization coefficient was set to 0.0002 and the weight updates used classical momentum (CM) in [4]. The initial momentum for the first 5 epochs was 0.5, the subsequent epochs used 0.9.

Since soft probabilities are needed to calculate the root mean square error, a version of SVM that computes class membership probability estimates was used. In the binary case, the probabilities are calibrated with Platt scaling [12], using logistic regression on the SVM's scores, fit by an additional cross-validation on the training data. For the multiclass cases, this procedure is extended as per [13]. These operations are very expensive and results have some inconsistencies, but they are not usually required in application.

## 3 Results

We start by comparing the results of our SVM model against our 1-layer neural network. Figure 2 compares the validation classification errors for every classifier. The all label refers to the percentage of training examples in which the model correctly predicts all classes. The classification errors were computed using the non-probabilistic version of SVM. The root square mean error of the on the other hand, was computed using the probabilistic version discussed earlier. An average RMSE of 0.1340 was obtained for the neural network and 0.1567 for the SVM.

The neural network outperforms the SVM in almost all classifiers, suggesting it is a viable model for this problem. Although SVM are a very different model in which the objective function has a dual representation that uses quadratic programming optimization, the algorithm is still very robust and widely used.

We now proceed to compare the impact of training with Nesterov's Accelerated Gradient (NAG), classical momentum (CM) and regular gradient descent. To keep computations short, we used our 1-layer network model and trained for 30 epochs with no early stopping. Weights are initialized using a Gaussian distribution with zero mean and variance 0.01, the same values are shared across the different methods. Figure 3 displays the importance of using momentum. NAG and CM can reach better values in fewer iterations. Interesting enough, CM seems to perform better at the end of training. In practice however, the model will start to overfit before that point, an early stopping will return similar validation errors.

Deep neural networks are finally brought into the analysis. Figure 4 compares them. Both approaches do better than the 1-layer neural network. The training and validation RMSE have gone down to 0.1105 and 0.1328 respectively, given by the model using unsupervised pretraining with RBMs, which performs slightly better.

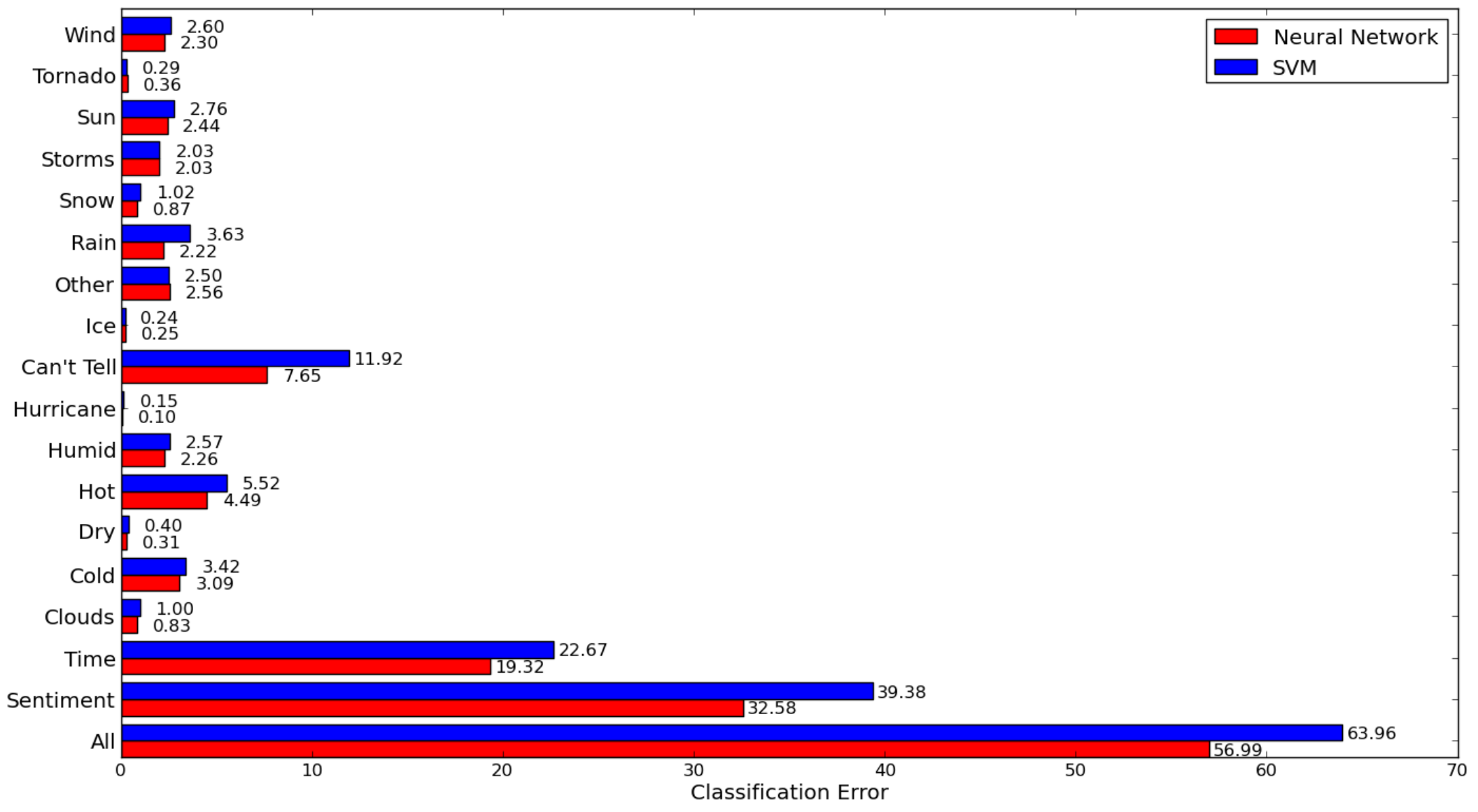

Figure 2: Comparison of the validation error rates for every classifier. Most important classifiers such as time of occurrence and sentiment are considerably outperformed by the neural network.

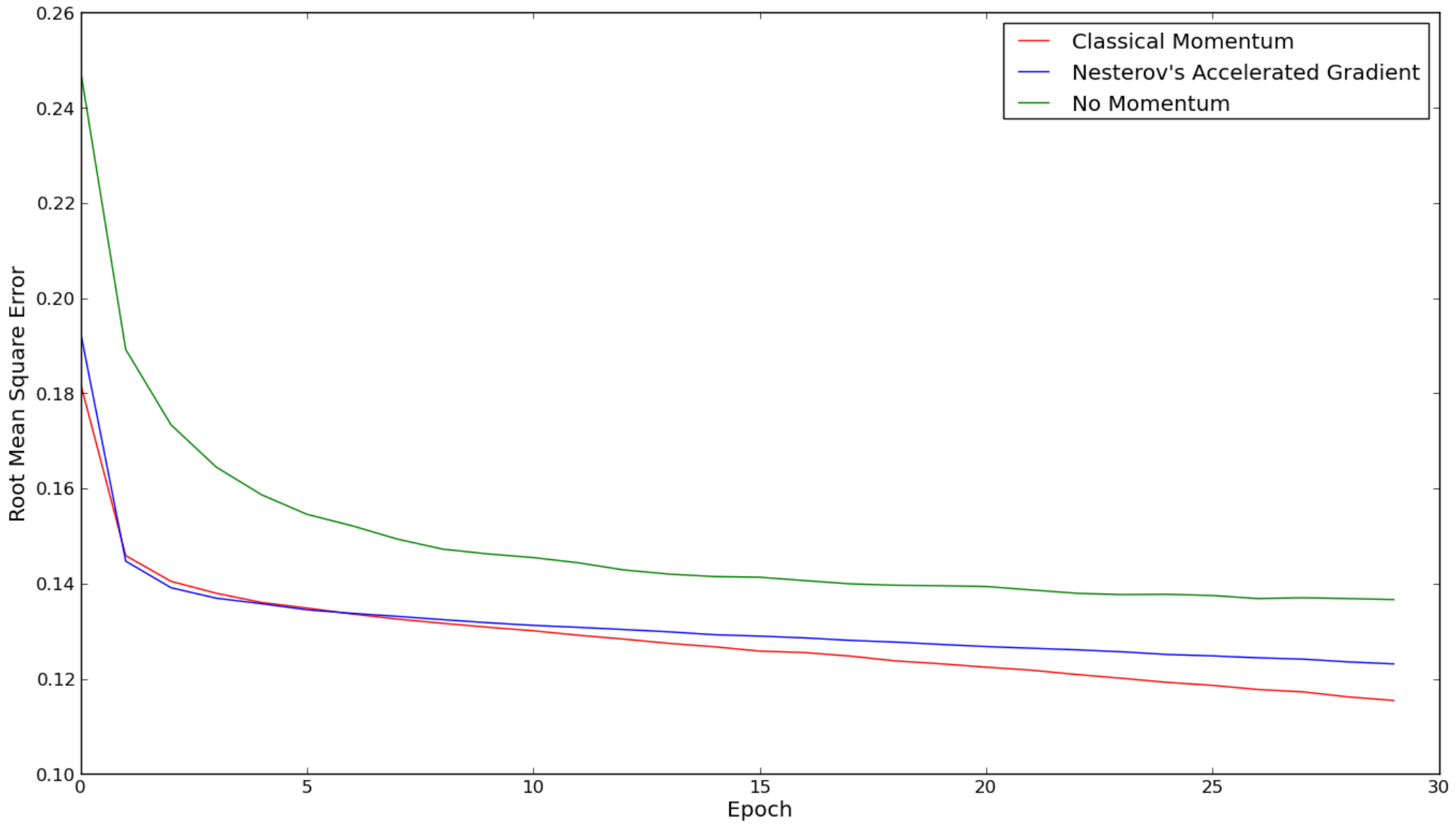

Figure 3: Training RMSE of a 1-layer neural network during 30 epochs under different weight update methods: Nesterov's Accelerated Gradient, classical momentum and regular gradient descent. NAG and CM get better results in less iterations than regular gradient descent.

Since the classification problem is part of a competition proposed by [5], we were able to compare results with those from other models. From 247 competitors sending an average of 14 entries each, our best model puts us in position 26 with a RMSE of 0.15638 on a test set. The best RMSE found in the competition is of 0.14314.

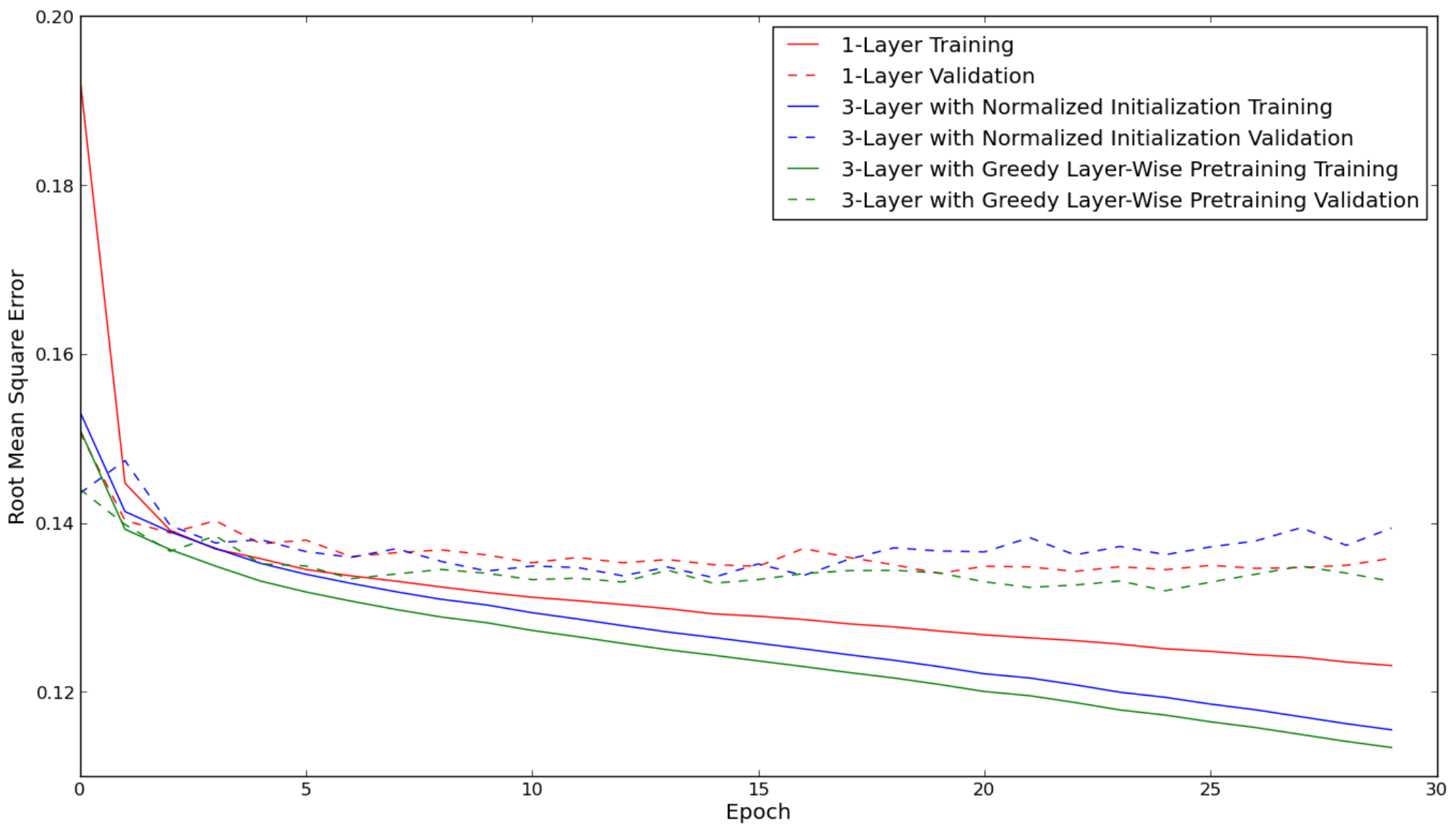

Figure 4: Training and validation RMSE of 3 models in which two 3-hidden-layer neural networks with normalized initialization and greedy layer-wise pretraining are able to outperform the single layer model.

## 4 Conclusions

It was shown that neural networks have at least a 6.97% improvement over non-linear SVMs. Performance gains are observed when using deep neural networks. DBN's greedy layer-wise pretraining still seems to give better results than other initialization methods. We conclude that the use of deep neural networks is a reasonable approach for this classification task and believe it can tackle other text classification problems as well.

Nonetheless, when data is abundant, pretraining weights can only do so much. Even with random initialization, after training for some epochs, units from upper layers would leave saturation and learning in the lower layers will start. That leaves us with other challenges, such as controlling overfitting and finding better minimum. To improve our results we suggest exploring techniques such as denoising autoencoders or dropout with Maxout Networks that will help take advantage of overfitting models [1]. Hessian Free optimization as proposed in [14] is a very interesting algorithm that employs second order methods.

Better results can also come from improving our feature extraction procedure. We suggest studying the effects of adding bigrams to the feature vectors. This would give words some context that could help with classification. For instance, keeping the adjective and noun together can have a great impact in sentiment analysis. The bigrams can undergo the same filtering and clustering procedures we designed.

### Acknowledgments

We thank Kevin Swersky for his insights, explanations and suggestions. Navdeep Jaitly for providing scaffolding code that helped with our implementation of neural networks, and Dr. Richard Zemel for teaching many of the concepts involved in this project.

### References

[1] Bengio, Y. Deep learning of representations: looking forward. *Technical report arXiv:1305.0445, Universite de Montreal, 2013*.

[2] Glorot, X., and Bengio Y. Understanding the difficulty of training deep feedforward neural networks. In *AISTATS, 2010*.

[3] Hinton, G. E., Osindero, S., and Teh, Y. A fast learning algorithm for deep belief nets. In *Neural Computation, 2006*.

[4] Sutskever I., Martens J., Dahl G., and Hinton G. On the importance of initialization and

momentum in deep learning. In *ICML, 2013*.

[5] CrowdFlower Open Data Library. Partly sunny with a chance of hashtags. *http://www.kaggle.com/c/crowdflower-weather-twitter*.

[6] Owoputi O., O'Connor B., Dyer C., Gimpel K., Schneider N., and Smith N. Improved part-of-speech tagging for online conversational text with word clusters. In *Proceedings of NAACL, 2013*.

[7] Bansal P. Hashtag segmentation. *http://github.com/piyushbansal/hashtag-segmentation*.

[8] Bird S., Klein E., and Loper E. Natural language processing with Python. *O'Reilly Media Inc., 2009*.

[9] Fellbaum C. WordNet: An Electronic Lexical Database. In *Cambridge, MA: MIT Press, 1998*.

[10] Mikolov T., Chen K., Corrado G., and Dean J. Efficient estimation of word representations in vector space. In *Proceedings of Workshop at ICLR, 2013*.

[11] Pedregosa et al. Scikit-learn: machine learning in Python. *JMLR, 2011*.

[12] Platt, J. Probabilities for support vector machines. In *Advances in large margin classifiers, MIT Press, 2000*.

[13] Wu, Lin and Weng, Probability estimates for multi-class classification by pairwise coupling. In *JMLR, 2004*.

[14] Martens, J. Deep learning via Hessian-free optimization. In *Proceedings of ICML, 2010*.